\documentclass[11pt,a4paper]{article}
\PassOptionsToPackage{table}{xcolor}

\usepackage{polyu-vclab}

\usepackage{lipsum}                
\usepackage{amsmath}
\usepackage{booktabs}
\usepackage{multirow}
\usepackage{wrapfig}
\usepackage{makecell}     
\usepackage{xcolor}       
\usepackage[numbers,sort&compress]{natbib}
\definecolor{lightpurple}{RGB}{230, 220, 235}
\definecolor{lightgreen}{RGB}{220, 230, 220}
\definecolor{lightpink}{RGB}{250, 240, 240}\providecommand{\ie}{\textit{i.e.}}
\providecommand{\eg}{\textit{e.g.}}

\setEyebrow{PolyU VCLab\,\textbullet\,ECCV 2026}

\setReportTag{Visual Computing Lab\,\textperiodcentered\,The Hong Kong Polytechnic University}

\papertitle{\centering\raisebox{-0.35em}{\includegraphics[height=1.2em]{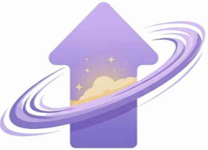}}\,Self-transcendence:\\ Is External Feature Guidance Indispensable for Accelerating Diffusion Transformer Training?}

\paperauthors{\centering%
  Lingchen Sun\equalmark\affilmark{1,2}\quad
  Rongyuan Wu\equalmark\affilmark{1,2}\quad
  Zhengqiang Zhang \affilmark{1,2}\quad
  Ruibin Li \affilmark{1}\quad
  Yujing Sun \affilmark{1,2}\quad
  Shuaizheng Liu \affilmark{1,2}\quad
  Lei Zhang\correspondmark\affilmark{1,2}%
}

\paperaffil{\centering%
  \affilmark{1}\,The Hong Kong Polytechnic University\quad
  \affilmark{2}\,OPPO Research Institute
}

\papernotes{\centering%
  \equalmark\,Equal contribution.\quad
  \correspondmark\,Corresponding author (\href{mailto:cslzhang@comp.polyu.edu.hk}{cslzhang@comp.polyu.edu.hk}).%
}

\paperbadges{%
  \vclabbadgesolid{Project Page}{https://yjsunnn.github.io/Self-Transcendence-project/}\;%
  \vclabbadgesolid{Code}{https://github.com/csslc/Self-Transcendence}\;%
}

\begin{document}

\maketitleVCLab

\begin{vclabAbstract}
\noindent\textbf{Abstract.}\;
Recent works such as REPA have shown that guiding diffusion models with external semantic features (\textit{eg.}, DINO) can significantly accelerate the training of diffusion transformers (DiTs). However, the use of pretrained external features as guidance signals introduces additional dependencies. We argue that DiTs actually have the power to guide the training of themselves, and propose \textbf{Self-Transcendence}, an effective method that achieves fast convergence using internal feature supervision only.
The desired internal guidance features should meet two requirements: \textit{structurally clean} to help shallow blocks separate noise from signal, and \textit{semantically discriminative} to help shallow layers learn effective representations.
With this consideration, we first align the DiT features with the clean VAE latent features, a native component of latent diffusion, for a short training phase (\textit{eg.}, 40 epochs) to improve their structural representations, then apply the classifier-free guidance to the intermediate features, enhancing their discriminative capability and semantic expressiveness.
These enriched internal features, learned entirely within the model, are used as supervision signals to guide a new DiT training from scratch.
Compared to existing self-contained methods, our approach achieves a significant performance boost. It can even surpass REPA, which uses the external DINO features as guidance, in both generation quality and convergence speed for both class-to-image and text-to-image generation tasks. Codes and models can be found at https://github.com/csslc/Self-Transcendence. 
\end{vclabAbstract}


\section{Introduction}

Diffusion models have emerged as a powerful framework for generative learning, achieving remarkable performance across a wide range of tasks, including image generation \cite{SD3, flux}, video synthesis \cite{wan, ma2025stepvideot2vtechnicalreportpractice, cogvideo, sora}, and multi-modal applications \cite{
begal, li2025llavaonevision, dai2023instructblip, qwen2025qwen25technicalreport}. Despite the great success, training diffusion transformers (DiTs) \cite{dit, sit} remains computationally intensive and suffers from slow convergence. Many methods \cite{repa, repae, disperse, tread, maskdit, LayerSync, sra, earlystop, zhu2024sdditunleashingpowerselfsupervised, wu2025representationentanglementgenerationtraining} have been developed to stabilize the DiT model training and accelerate the convergence process. Recent studies \cite{repa, disperse, singh2025matters} have highlighted the crucial role of meaningful intermediate representations in both improving training efficiency and enhancing generative capability.

To enrich feature representations, several representation learning strategies have been proposed, including masked training \cite{gao2023masked, maskdit, mdtv2, sddit}, contrastive learning \cite{disperse}, and representation alignment \cite{repa, repae, reed}.
Among them, the pioneering work REPA \cite{repa} introduces an effective regularization strategy to align DiT features with external vision encoders such as DINO \cite{dinov2}, significantly accelerating model training and improving generation performance. 
However, this success is highly dependent on external networks and introduces additional dependencies.

To eliminate the reliance on external supervision, recent works \cite{disperse, LayerSync, sra} have explored self-contained alternatives. Dispersive Loss \cite{disperse} introduces a plug-and-play regularizer that encourages feature dispersion without requiring pre-training or auxiliary data. SRA \cite{sra} and LayerSync \cite{LayerSync} instead leverage the discriminative features in deeper layers during training to guide the learning of shallower layers.
Specifically, SRA employs the EMA model as a teacher during training and performs layer-wise distillation across different noise levels. LayerSync introduces a lightweight synchronization mechanism to align semantically richer and weaker layers. Both of them eliminate the need for external feature extractors.
However, their performance lags behind externally-guided approaches such as REPA \cite{repa}. 
Therefore, a critical question arises: \textit{Is external feature guidance indispensable for accelerating diffusion transformer training?}

\begin{figure*}
	\centering 
	\includegraphics[scale=0.55]{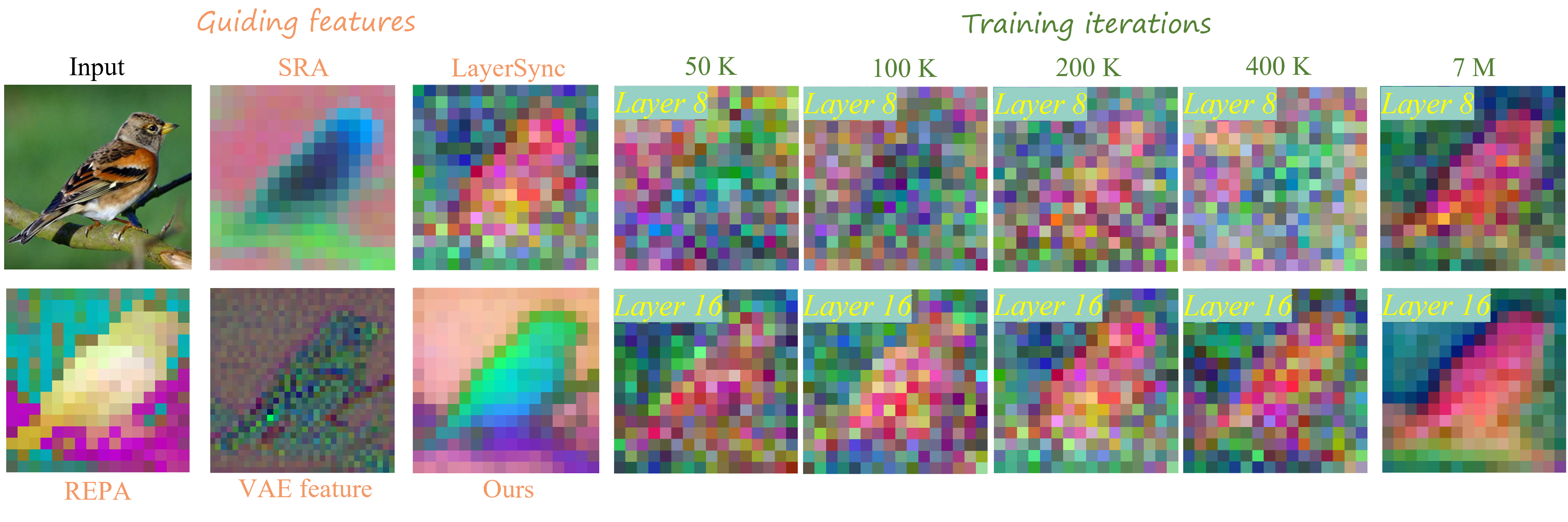}
	\caption{\textbf{Left}: Comparison of guiding features used in different methods. Our self-contained approach yields more structured and semantically meaningful features than SRA \cite{sra} and LayerSync \cite{LayerSync}, comparable to the external DINO \cite{dinov2} features used in REPA \cite{repa}.
    \textbf{Right}: PCA visualization \cite{abdi2010pca} of latent features from both shallow (layer 8) and deeper (layer 16) blocks of SiT with $t=0.6$ during training. Both layers progressively learn clean and discriminative representations, while the shallow layer learns at a slower pace compared to the deeper one.}
	\label{intro}
\end{figure*}

In this work, we aim to answer this question. We attribute the performance gap to the limited quality of internally generated guidance signals, especially in the early training stages.  
To serve as effective guidance signals, the features should satisfy two criteria: they should be \textit{structurally clean} to help shallow blocks disentangle noise from signal, and \textit{semantically discriminative} to enable shallow layers to learn informative representations \cite{singh2025matters}.
As shown on the left side of Fig.~\ref{intro}, when synthesizing an image of a bird, REPA (with DINO features) highlights semantically meaningful regions. By contrast, SRA/LayerSync rely on immature internal features and are semantically less expressive, making them ineffective for guiding shallow layers. 
Actually, the DiT architecture has the potential to provide useful semantic guidance \cite{lee2024dmp}, but this potential has not been fully unleashed.
In the right side of Fig.~\ref{intro}, we visualize the latent DiT features of a shallow block (layer 8) and a deeper block (layer 16) using PCA \cite{abdi2010pca} across training. We see that while both layers gradually become more discriminative, the shallow block evolves much more slowly. This indicates that the convergence of DiT is mainly bottlenecked by the learning of clean and semantically rich features in shallow layers, where better guidance signals are important.

Motivated by these observations, we propose \textbf{Self-Transcendence}, a fully self-guided training strategy that can surpass REPA-level performance but without external supervision. Self-Transcendence guides DiT training in two stages. The first stage performs \textbf{VAE Structure Guidance}. 
As shown on the left side of Fig. \ref{intro}, the guiding features of SRA and LayerSync are updated along with training and tend to be noisy and unstable in the early stages. To address this, we directly use the clean VAE latent features within the standard latent diffusion framework to guide the model building structured internal representations.
However, while providing a good starting point, the semantic richness of VAE features is limited. Therefore, we perform a second stage of \textbf{Self-guided Representation Alignment}.
After a warm-up phase, we extract intermediate features from the deeper layer of the partially trained model and apply classifier-free guidance (CFG) to help the guidance feature obtain stronger internal semantics.

This two-stage strategy not only enjoys the benefits of the standard latent diffusion framework (\ie, VAE feature), but is also entirely self-contained, achieving self-transcendence for training DiT models.
Training the guiding model adds modest overhead but substantially speeds up subsequent DiT training. Unlike prior methods that rely on external vision encoders \cite{repa, reed}, 
our method shows that the internal features of the DiT model itself can serve as an effective guidance. The proposed approach does not require external data and model and is overall more cost-effective. 

Our contributions are summarized as follows:
\begin{itemize}
    \item We introduce Self-Transcendence, a fully self-guided training framework to improve DiT model training without relying on external features.
    \item We propose a two-stage pipeline to obtain a semantically richer internal representation, which first aligns shallow-layer with VAE latent features, and then enhances intermediate features using CFG.
    \item We show that our method achieves even better performance than REPA in both convergence and generation quality, but without using external data and external features.
    
\end{itemize}

\section{Related Work}
\label{sec:related}
\textbf{Architectures of Diffusion Models.}
Early diffusion models typically adopt a U-Net backbone \cite{ldm}, consisting of convolution and attention layers \cite{attention}. Recently, Diffusion Transformers (DiTs) \cite{dit} replace U-Net with pure transformer blocks, following the design of Vision Transformers \cite{vit}.
To further enhance DiT, Scalable Interpolant Transformers (SiT) \cite{sit} introduce a more flexible interpolant framework to generalize the diffusion process, systematically exploring the design choices of time discretization, prediction targets, interpolant types, and sampling strategies.
LightningDiT \cite{vavae} pushes DiT to its performance limits by incorporating a range of training and architectural optimizations, such as RMSNorm \cite{zhang-sennrich-neurips19}, SwiGLU \cite{shazeer2020gluvariantsimprovetransformer}, and RoPE \cite{su2021roformer}, enabling faster convergence and more efficient inference. Other efforts investigate architectural improvements such as U-shaped transformer designs \cite{udit}, dynamic computation adjustment \cite{zhao2024dynamic}, mixture-of-experts \cite{anonymous2025densemoe}, linear attention mechanisms \cite{xie2024sana}, and decoupled transformer design \cite{wang2025ddt} to further boost the scalability and efficiency of diffusion models.
Our proposed Self-Transcendence presents a new training strategy for the DiT family, guiding the model itself to converge faster.

\noindent\textbf{Acceleration of DiT Training.}
Recent studies \cite{repa,repae} highlight the importance of semantically meaningful representations to improve training efficiency and generation quality. 
MaskDiT \cite{maskdit} accelerates DiT training by randomly masking 50\% of input patches, encouraging efficient learning of informative features.
MAETok \cite{chen2025maetok} applies the masking strategy to tokenizer training, improving diffusion models by learning a semantically structured latent space.
RCG \cite{rcg} learns a generative model that generates semantic representations extracted by a self-supervised encoder, using them as conditions for image generation.
REPA \cite{repa} introduces a representation alignment loss that aligns internal features with pretrained image embeddings \cite{dinov2}, significantly boosting training speed and generation performance.
The following works explore the use of pretrained vision encoders to provide richer external supervision.
U-REPA \cite{urepa} extends REPA to the U-Net architecture.
VA-VAE \cite{vavae} addresses the optimization bottleneck in latent diffusion by aligning the latent space of a tokenizer with pretrained vision foundation models, enabling faster convergence and better generation quality in high-dimensional settings.

In contrast, some recent works aim to accelerate training without pretrained vision encoders.
Dispersive Loss \cite{disperse} promotes diverse internal representations during diffusion training without external feature guidance.
SRA \cite{sra} and LayerSync \cite{LayerSync} guide shallow layers using semantically richer internal features. However, these methods still lag behind REPA in terms of performance.
To bridge this gap, we propose Self-Transcendence, which leverages DiT model’s own representations as a substitute for external features.

\section{Method}
\label{sec:method}

\begin{figure*}
	\centering 
	\includegraphics[scale=0.93]{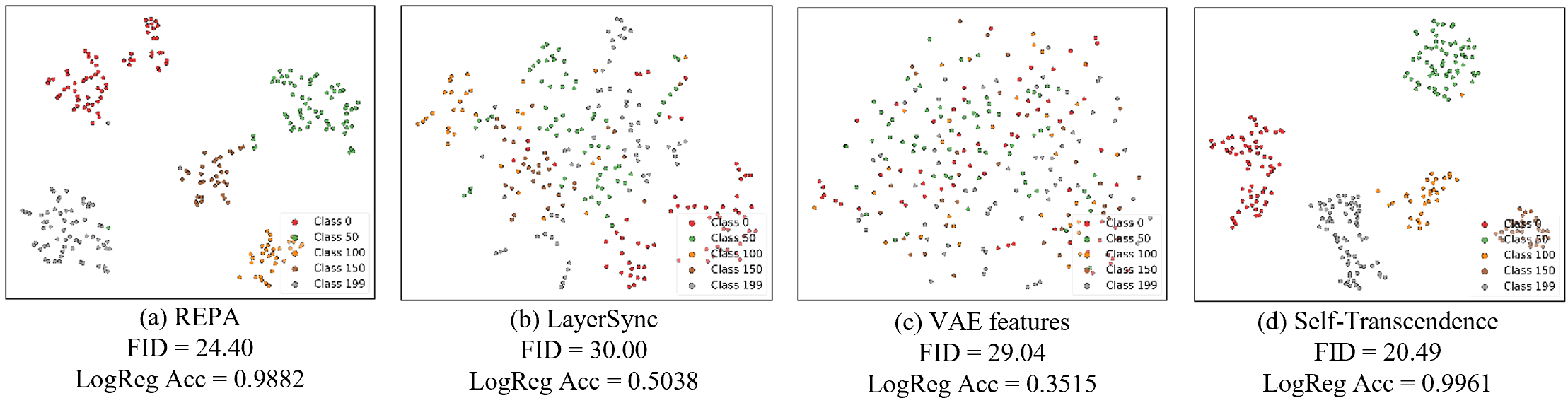}
	\caption{{t-SNE visualizations of the guiding features extracted from (a) REPA \cite{repa}, (b) LayerSync \cite{LayerSync}, (c) VAE features, and (d) our Self-Transcendence with $t=0.4$ in the 200K iteration of SiT-XL/2. Different colors represent different classes. Our internal guiding features demonstrate superior class separability, comparable to REPA.}
    }
	\label{motivation}
\end{figure*}

\subsection{Motivation and Framework Overview}
\label{subsec:motivation}
With the widespread adoption of DiTs \cite{dit, sit, vavae}, accelerating their training has become an important research topic \cite{udit, vavae, maskdit, wang2025ddt, repa}. Generally speaking, shallow blocks of DiT models are responsible for discriminative tasks, \textit{ie.}, separating clean latent states from noise in the given noisy input, while deeper blocks focus on refining details based on the shallow representations.
However, as training progresses, a challenge emerges: shallow layers become slow to learn discriminative features (see Fig. \ref{intro}) due to the long gradient propagation path. This observation has motivated a line of research \cite{repa, wang2025ddt, repae, LayerSync, sra, disperse} to explore how to train shallow blocks with better discriminative representations.

One promising approach is to introduce guiding signals to supervise the learning of shallow features. REPA \cite{repa} initiates this research by using external DINO features \cite{dinov2} to guide shallow DiT layers. DINO is a self-supervised vision encoder that learns powerful semantic representations. As shown in Fig. \ref{motivation}(a), DINO embeddings form clear clusters, indicating strong semantic separability. These features boost the learning of shallow DiT layers and significantly accelerate DiT training.
However, relying on external DINO features introduces new dependencies: it requires extensive pre-training with external data, which may not always be feasible and desirable. Recent works have begun to explore whether diffusion models can achieve self-acceleration. For example, Layersync and SRA \cite{LayerSync, sra} use deeper layer features to guide shallower layer features. However, these features lack stable structural guidance and  semantic discriminability, leading to weaker performance, as shown in Fig. \ref{motivation}(b). 

In addition to features with discriminative power, we find that features with clean structures \cite{singh2025matters} in the latent diffusion pipeline, can also be important for guiding features. Although VAE lacks strong discriminative power,
its latent space is clean and structured. Surprisingly, VAE features can accelerate DiT training to a level comparable to LayerSync, as shown in Fig. \ref{motivation}(c), suggesting that structured features are effective for providing effective guidance, even without high discriminability. 

\begin{figure*}
	\centering 
	\includegraphics[scale=0.5]{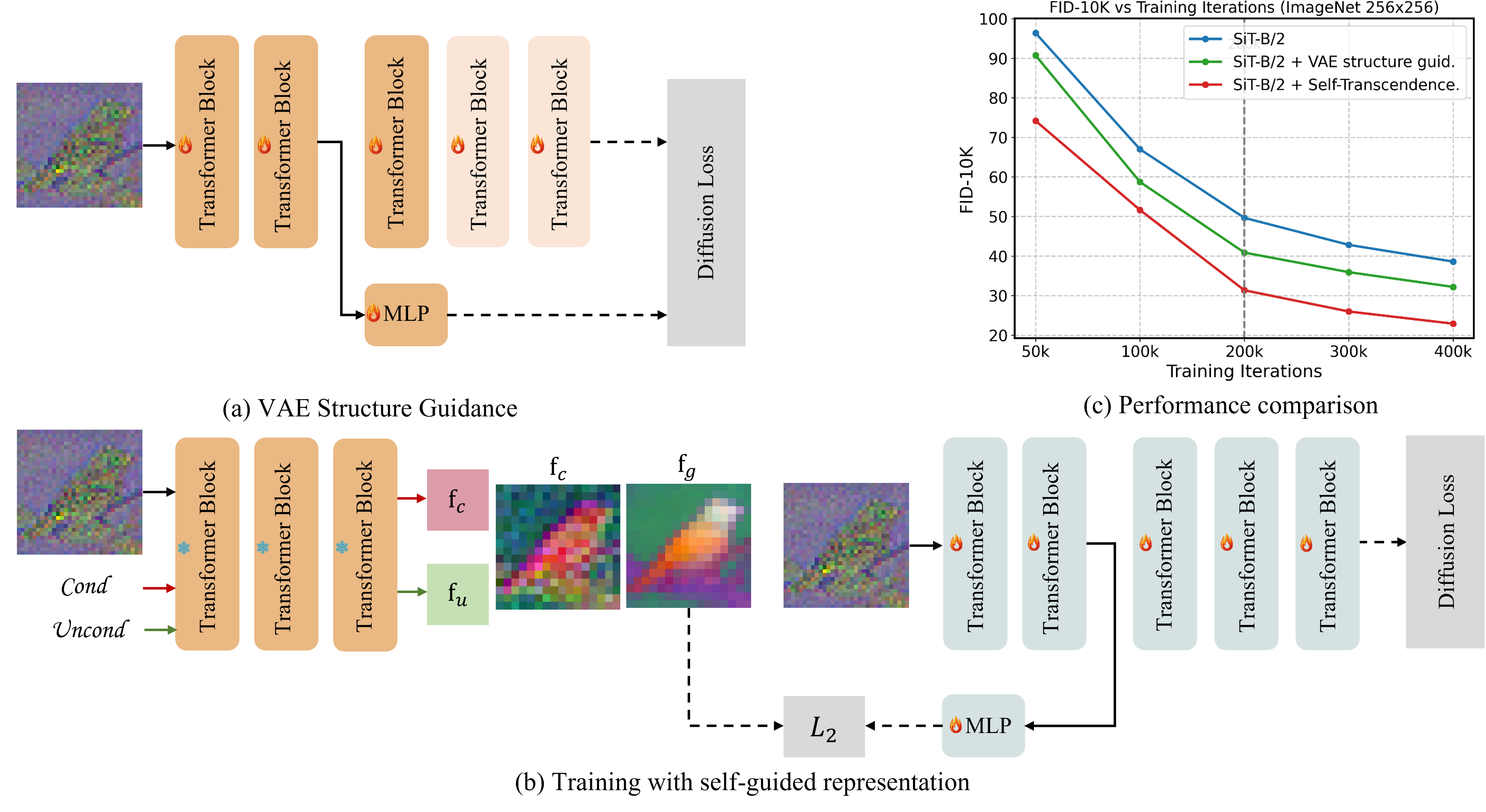}
	\caption{The framework of our proposed \textbf{Self-Transcendence} approach. The spark icon indicates that the parameters of this layer are trainable, while the snowflake icon indicates that they are frozen.}
	\label{framework}
\end{figure*}

Based on the above findings, we argue that effective guiding features should meet two criteria:  (1) \textit{they should have a clean structure so that they can effectively help shallow layers distinguish noise from signal}, and (2) \textit{they should be semantically discriminative so that they can help shallow layers to learn effective representations.} 
Several works \cite{lee2024dmp, li2023your, clark2023texttoimage, meng2024not} have pointed out that diffusion models can be leveraged for various discriminative tasks, such as classification. This further motivates us to think whether we can find a discriminative and semantically enriched feature representation inside the DiT model to guide the training of itself, achieving self-transcendence.

With these considerations, we propose a new DiT training acceleration framework, namely \textbf{Self-Transcendence}, as illustrated in Fig. \ref{framework}. Unlike one-stage methods such as SRA \cite{sra} and LayerSync \cite{LayerSync}, which update the guiding feature during model training, we adopt a two-stage design, which first warms up a meaningful representation and then uses it to guide a new model learning, for a more stable guiding process. Specifically, in the first stage, we use clean VAE features as guidance to help the model learn a structurally clean representation to distinguish useful information from noise in shallow layers (see Fig. \ref{framework}(a)). After a certain number of iterations, we freeze this model and use its representation as a fixed teacher.
To enhance the semantic expression of the features, we then perform a self-guided representation to better align with the target conditions (see Fig. \ref{framework}(b)).
Compared to REPA that uses external DINO features, our method replaces external features with features from a partially trained internal model, which demonstrates highly competitive guiding performance.
Although this warm-up introduces a little extra cost, it significantly accelerates the subsequent model training and improves performance (see Fig. \ref{framework}(c)). (A one-stage comparison is provided in the \textcolor{polyured}{\textbf{Appendix}}.)

\subsection{VAE Structure Guidance}

Standard diffusion models are trained with a single denoising loss imposed on the output, which often leads to slow learning in shallow layers.
To improve this, we introduce an auxiliary VAE structure guidance loss to intermediate layers, as shown in Fig. \ref{framework}(a). Specifically, at the $n$-th layer, we extract the intermediate feature $\mathbf{f}_n$, pass it through a lightweight multilayer perceptron (MLP), and align it with the ground-truth VAE latent $\mathbf{z}$ using an $L_2$ loss, as shown below: 
\begin{equation}
\mathcal{L}_{\text{VAE-guide}} = \left\| \text{MLP}(\mathbf{f}_n) - \mathbf{z} \right\|_2^2.
\label{vae}
\end{equation}
The latent representation of the VAE provides a clean structural prior that helps the shallow layers distinguish meaningful signals from noisy inputs.

This alignment can be regarded as an additional diffusion loss constrained only in shallow layers, without adding extra computational resources. Therefore, the shallow blocks are facilitated to perceive structural information and speed up model convergence. As shown in Fig. \ref{framework}(c), this simple alignment alone can already accelerate the learning of DiT, even obtaining better performance than the existing self-contained methods (please refer to Table \ref{tab:400k}).
However, while VAE structure guidance improves structural learning, it is not as effective as external features (\eg, DINO) for semantic alignment. To address this, we propose a self-guided representation in the following section.

\subsection{Self-guided Representation}
During the training of a diffusion model, the internal features gradually become more discriminative. Previous works have shown that features from deeper layers often capture stronger semantic information \cite{repa,LayerSync,sra}. Therefore, we use deep-layer features as guiding signals to improve the training of a new model, as shown in Fig. \ref{framework}(b). However, as shown in Figs. \ref{motivation}(a) and (b), even with 200K training steps, the semantic richness of these features still lags behind that of pretrained external vision encoders like DINO.

To address this, we draw inspiration from Classifier-Free Guidance (CFG) \cite{cfg}, a technique widely used in conditional diffusion models. CFG improves the alignment between generated samples and the conditioning input without requiring an external classifier. During training, the model randomly removes the condition to learn both conditional and unconditional denoising. In the inference stage, it combines predictions from both modes using a guidance scale:
\begin{equation}
\mathbf{o}_{g} = \mathbf{o}_{u} + \omega \cdot (\mathbf{o}_{c} - \mathbf{o}_{u}),
\label{guide}
\end{equation}
where $\mathbf{o}_{c}$ and $\mathbf{o}_{u}$ denote diffusion predictions with the condition and uncondition inputs. Increasing $\omega$ strengthens the influence of the conditional signal, generating images that better match the desired condition.

Building upon this idea, we extend the CFG from the output space to the feature space. We extract both conditional and unconditional features from the same layer under the same input. Then the two features are combined using a guidance scale, as shown in Eq. (\ref{guide_fea}):
\begin{equation}
\mathbf{f}_{g} = \mathbf{f}_{u} + \omega \cdot (\mathbf{f}_{c} - \mathbf{f}_{u}),
\label{guide_fea}
\end{equation}
where $\mathbf{f}_{c}$ and $\mathbf{f}_{u}$ are the conditional and unconditional features, respectively.
Eq. (\ref{guide_fea}) encourages internal representations to align more closely with the desired semantics.
As illustrated in Fig. \ref{framework}(b) and Fig. \ref{motivation}(d), this feature-level guidance highlights the semantic region and significantly improves the discriminative separability of deep features $\mathbf{f}_{g}$ compared to their original counterparts $\mathbf{f}_{c}$.

We then use the enriched feature $\mathbf{f}_g$ to guide the shallower layers together with the standard diffusion loss, \ie, $\mathcal{L} = \mathcal{L}_{diff} + \lambda_{guide} \times \mathcal{L}_{guide}$, as shown in Fig. \ref{framework}(b). Specifically, intermediate features $f_m$ of the trained DiT are passed through a lightweight MLP (whose architecture is detailed in the \textcolor{polyured}{\textbf{Appendix}}), and an $L_2$ loss is applied to align them, as shown below:
\begin{equation}
\mathcal{L}_{guide} = \left\| \text{MLP}(\mathbf{f}_m) - \mathbf{f}_{g} \right\|_2^2.
\label{final_loss}
\end{equation}

The self-guided representation explores the intrinsic discriminative ability of diffusion models by applying CFG on deep features to improve the discriminative power of the shallow layers in DiT, further accelerating DiT training, as illustrated in Fig. \ref{framework}(c).

\section{Experiments}
\label{sec:Experiment}

\subsection{Experimental Setup}

\textbf{Implementation Details}.
We conduct experiments using two baseline models: SiT \cite{sit} with a patch size of 2, and LightningDiT \cite{vavae} with a patch size of 1. To ensure a fair comparison, we follow the training and inference settings of each baseline.
For the proposed Self-Transcendence method, several components should be determined: the guiding model, the guidance scale $\omega$ in the self-guided representation, loss weight $\lambda_{guide}$, as well as the guided and guiding layers. For all baselines, we use the model trained with the VAE structure guidance at 40 epochs as the guiding model. Note that the guiding model shares the same architecture as the baseline model. We set the guidance scale to $\omega = 30.0$ and $\omega = 10.0$ for the SiT and LightningDiT backbones, respectively. Suppose that the total number of Transformer blocks is $n$, we select the guided layer as $n/2$ and the guiding layer as $2n/3$. The ablation studies on these parameters are given in Sec. \ref{sec:ablation}.

We apply an early stop strategy during training. The self-guided representation loss is only used during the early iterations (20 epochs for base models, 10 epochs for larger models from our experimental study). Then, we remove the self-guided loss and only optimize the diffusion loss. This design aims to improve the training of shallow layers, which lack semantic structures. However, over-training the shallow layers can make the training of deeper layers unstable. Similar phenomena \cite{earlystop} have been observed in REPA. Further discussion is provided in  the \textcolor{polyured}{\textbf{Appendix}}.

\noindent\textbf{Evaluation Metrics}.
We employ commonly used evaluation metrics \cite{sit, repa}, including the Fréchet Inception Distance (FID) \cite{fid}, sFID \cite{sfid}, inception score (IS) \cite{is}, precision, and recall \cite{precision}.
Metrics are computed on 50k samples for Tables \ref{tab:400k} and \ref{tab:cfg_compare}, and on 10k samples for Tables \ref{tab:ablation_com} and \ref{tab:layers-scale}.

\noindent\textbf{Compared Methods}.
We compare with different acceleration methods: REPA \cite{repa}, Disperse Loss \cite{disperse}, SRA \cite{sra}, and LayerSync \cite{LayerSync}. Various latent diffusion
models are also compared. (1) U-Net backbone: LDM \cite{ldm}. (2) Hybrid 
Transformer and U-Net backbone: U-ViT-H/2 \cite{u-vit} and MDTv2-XL/2 \cite{mdtv2}. (3) Transformer backbone: MaskDiT \cite{maskdit}, SD-DiT \cite{sddit}, DiT-XL/2 \cite{dit}, and SiT-XL/2 \cite{sit}.

\subsection{Main Results}
\textbf{Comparison with Existing Acceleration Methods.}
Table \ref{tab:400k} shows the comparison between our proposed VAE structure guidance and Self-Transcendence methods with other acceleration approaches. The following observations can be made.
\textbf{(1)} Firstly, VAE structure guidance alone achieves competitive results with existing self-contained methods such as Disperse Loss and LayerSync. This is because semantic structure is mainly learned during the early training stages. However, Disperse Loss and LayerSync fail to provide strong and stable guidance in this phase. In contrast, our guiding features are derived from a fixed VAE,  making it more stable and easier to guide the learning process. 
\textbf{(2)} Secondly, our proposed Self-Transcendence method significantly outperforms all other self-contained techniques. On SiT-XL/2, our method achieves 7.51 FID with only 80 epochs, outperforming the LayerSync trained for 200 epochs (8.80 FID). In addition, Self-Transcendence achieves results comparable to or even better than REPA, which uses external DINO features. 
\textbf{(3)} Thirdly, our approach also brings substantial improvements to LightningDiT, showing that it can be generalized to 
\begin{wraptable}{r}{0.6\textwidth}
\vspace*{-2pt} 
\caption{Comparisons with different acceleration methods based on the vanilla SiTs and LightningDiTs on ImageNet 256$\times$256. CFG is not used. $\downarrow$ denotes that lower values are better.}
\label{tab:400k}

\resizebox{\linewidth}{!}{
\begin{tabular}{@{}lccc@{}}
\toprule
\textbf{Model} & \textbf{\#Params} & \textbf{Epochs} & \textbf{FID$\downarrow$} \\
\midrule
SiT-B/2 & 130M & 120 & 31.45 \\
SiT-B/2 & 130M & 80 & 36.14 \\
\quad + REPA & 130M & 80 & 24.40 \\
\quad + Disperse Loss & 130M & 80 & 32.45 \\
\quad + LayerSync & 130M & 80 & 30.00 \\
\rowcolor{lightpurple} \quad + VAE structure guidance (Ours) & 130M & 80 & 29.04 \\
 \rowcolor{lightpurple} \quad + Self-Transcendence (Ours)& 130M & 80 & \textbf{20.49} \\
\midrule
SiT-L/2 & 458M & 80 & 21.41 \\
\quad + REPA  & 458M & 80 & 9.70 \\
\quad + Disperse Loss & 458M & 80 & 16.68 \\
\quad + LayerSync & 458M & 80 & 14.83 \\
\rowcolor{lightpurple} \quad + VAE structure guidance (Ours) & {458M} & {80} & {14.61} \\
\rowcolor{lightpurple} \quad + Self-Transcendence (Ours) & {458M} & {80} &
\textbf{8.74}  \\
\midrule
SiT-XL/2 & 675M & 800 & 8.30 \\
SiT-XL/2 & 675M & 120 & 14.74 \\
SiT-XL/2 & 675M & 80 & 17.63 \\
\quad + REPA  & 675M & 80 & 7.90 \\
\quad + Disperse Loss & 675M & 200 & 10.64 \\
\quad + LayerSync & 675M & 200 & 8.80 \\
 \rowcolor{lightpurple} \quad + {VAE structure guidance (Ours) }& {675M} & {80} & {12.25} \\
 \rowcolor{lightpurple} \quad {+ Self-Transcendence (Ours)} & {675M} & {80} &
\textbf{7.51} \\
\midrule
LightningDiT-B/1 (w. VAVAE)& 130M & 64 & 15.94 \\
\rowcolor{lightpurple} \quad {+ VAE structure guidance (Ours)} & {130M} & {64} & {15.87} \\
\rowcolor{lightpurple} \quad {+ Self-Transcendence (Ours)} & {130M} & {64} & \textbf{14.03} \\
\midrule
LightningDiT-XL/1 (w. VAVAE)& 675M & 64 & 5.30 \\
\quad + REPA & 675M & 64 & 4.09 \\
\rowcolor{lightpurple} \quad {+ VAE structure guidance (Ours)} & {675M }& {64 }& {5.07} \\
\rowcolor{lightpurple} \quad {+ Self-Transcendence (Ours)} & {675M} & {64} &
\textbf{3.55} \\

\bottomrule
\end{tabular}}

  \vspace*{-2pt} 
\end{wraptable}
different backbones (SiT and DiT) and different VAE latent spaces (SD-VAE \cite{ldm} and VAVAE \cite{vavae}). Notably, on LightningDiT-XL/1, Self-Transcendence achieves an FID of 3.55 with only 64 epochs of training.

Finally, although our method requires to train a guiding model using VAE structure guidance, this overhead is minimal and worthwhile. As shown in SiT-B/2 and SiT-XL/2, it outperforms the longer-trained baselines by a large margin. This demonstrates that a small cost in warm-up training can lead to significant acceleration and performance gains. Moreover, compared to the widely used guiding model (DINO), the training for our guiding model is much easier and more efficient without using external data. 

\begin{figure*}[t]
	\centering 
	\includegraphics[scale=0.9]{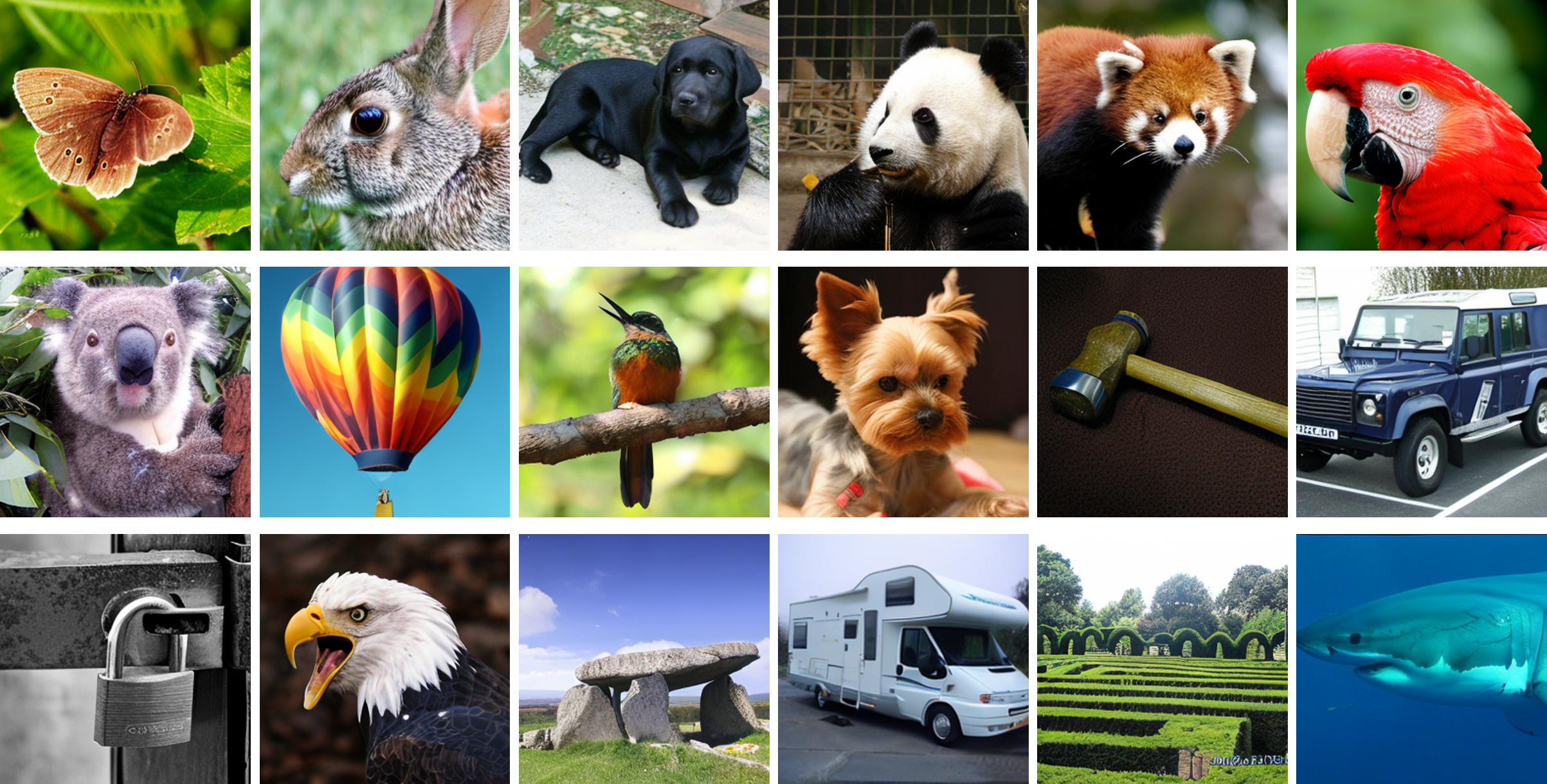}
	\caption{Examples of generated images on ImageNet $256 \times 256$ of our proposed Self-Transcendence method. We use classifier-free guidance with scale 4.0.
    }
	\label{generated256}
\end{figure*}

\noindent\textbf{Scalability.} 
We evaluate the proposed Self-Transcendence across different model sizes. As shown in the Table~\ref{tab:400k}, our method consistently accelerates training at all scales. 
Notably, the performance gain becomes larger as the model size increases. For example, on SiT/B-2, the FID improves from 36.14 (baseline) to 20.49 with a 43.3\% relative reduction. On SiT/XL-2, it improves from 17.63 to 7.51, with a 57.4\% reduction. This clearly demonstrates the scalability of our method.
In addition, our guiding model shares the same backbone as the generation model, which may benefit from larger generation models. 
We provide qualitative results of SiT-XL/2 on ImageNet at resolutions of 256 using the proposed Self-Transcendence 
method in Fig. \ref{generated256}, showcasing its ability to generate realistic structures and textures across diverse semantic categories. For animals, such as dog, panda, and bird, our method captures fine-grained details such as fur texture, feather 
patterns, and facial features with high fidelity.
In natural scenes, such as the stone, our trained models generate depth and organic shapes 
with coherence and realism. 
For man-made and structured objects such as cars, the generated images exhibit accurate geometry and sharp edges.
To further evaluate the scalability of our method, we conduct experiments on ImageNet at a \textbf{resolution of 512 $\times$ 512}. Experimental details can be found in \textcolor{polyured}{\textbf{Appendix}.}

\begin{wraptable}{r}{0.6\textwidth}
\vspace*{-2pt} 
\caption{{Comparisons of the proposed self-transcendence with REPA in the Text-to-Image task.  $\downarrow$ and $\uparrow$ indicate the lower and higher values are better, respectively.}}
\label{tab:t2i}
\centering
\scriptsize
\begin{tabular}{lccc}
\toprule
\textbf{Model} & \textbf{iters} & \textbf{FID$\downarrow$}  & \textbf{IS$\uparrow$}   \\
\midrule
MMDiT & 150k & 6.23 & 30.74 \\
\midrule
\quad + REPA & 150k & 4.90 & 32.55 \\
\midrule
\rowcolor{lightpurple} \quad + Self-Transcendence & 150k & \textbf{4.56} & \textbf{33.08} \\
\bottomrule
\end{tabular}

  \vspace*{-2pt} 
\end{wraptable}

\begin{figure*}[t]
	\centering 
	\includegraphics[scale=0.70]{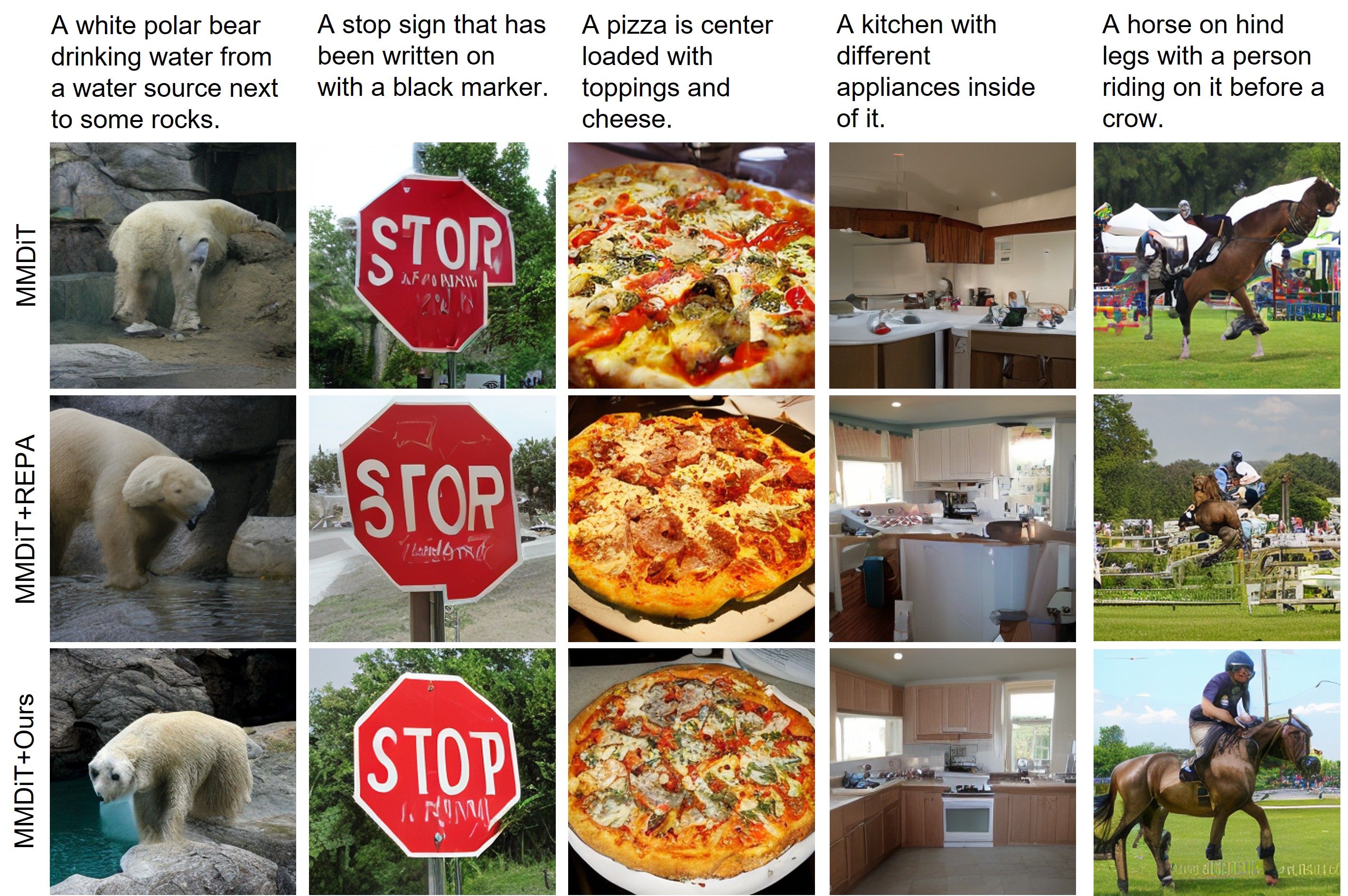}
   \captionsetup{belowskip=-1.5em}   
	\caption{Qualitative comparison on text-to-image generation (MS-COCO).
    }
	\label{t2i}
\end{figure*}

We also conduct text-to-image (T2I) generation experiments using the same setting (including training dataset evaluation protocol) as REPA \cite{repa}. The results are shown in Table \ref{tab:t2i}. We can see that Self-Transcendence also works well on the more complex T2I generation task, showing better generation scores than REPA (FID: 4.56 vs. 4.90; IS: 33.08 vs. 32.55). Some visual comparisons of the generated images are shown in Fig. \ref{t2i}. We can see that REPA improves over original MMDiT with cleaner structure and fewer artifacts. Our method 
further improves prompt alignment and visual performance, yielding more realistic results with faithful structure and illumination.

\begin{figure*}[t]
	\centering 
	\includegraphics[scale=0.60]{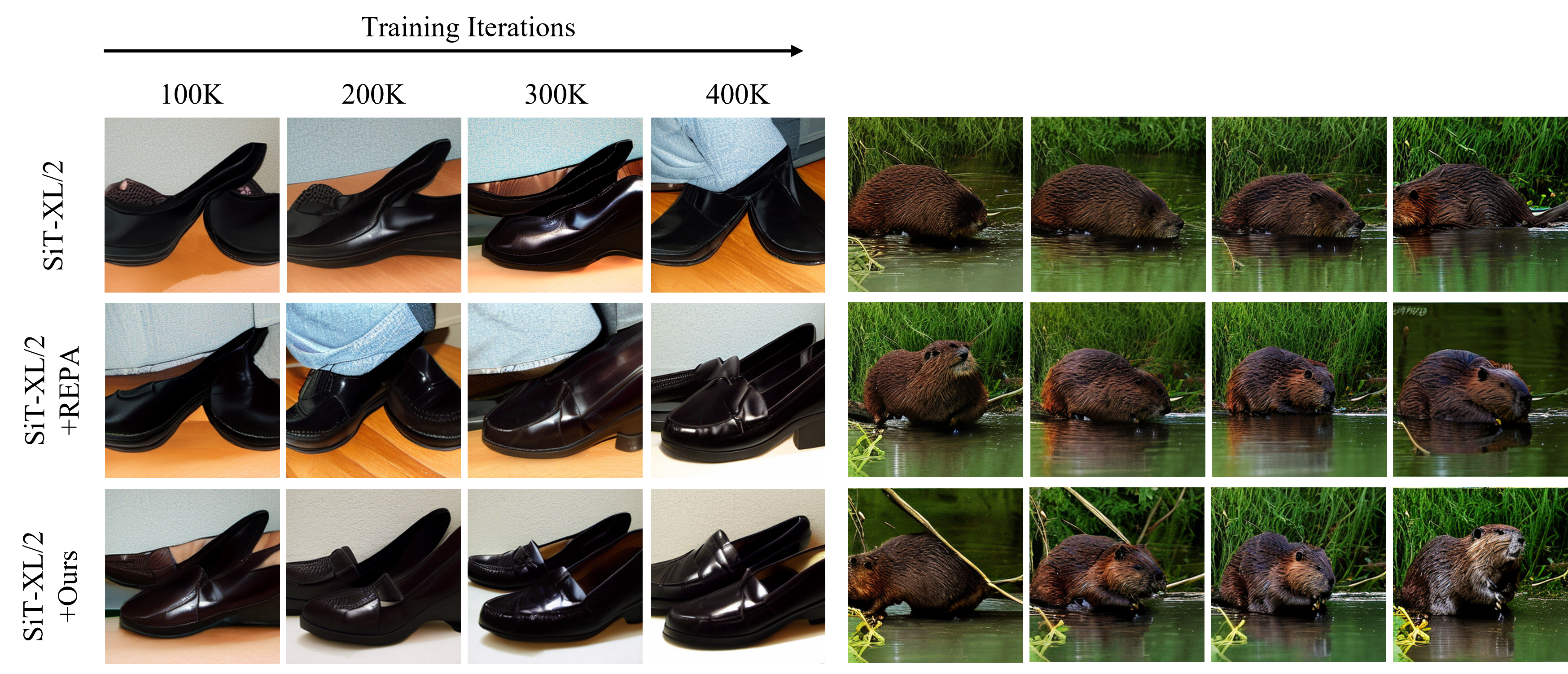}
	\caption{Visual comparison of generated samples from SiT-XL/2 models at different training iterations. For all models, we apply the same seed, noise, and sampling strategy with a CFG scale of 4.0.}
	\label{visual_com}
\end{figure*}

\noindent\textbf{Visualization of Training Process}.
We compare the vanilla SiT model, REPA-enhanced model, and our trained model across 100K to 400K iterations on two ImageNet classes, as illustrated in Fig.~\ref{visual_com}. Both our model and REPA show faster convergence than the vanilla SiT model, and our method tends to obtain better structural generation. For example, in the shoe class, our method generates more realistic shapes and consistent textures earlier in training. By 400K iterations, our model yields sharper contours and finer details, indicating improved sample quality and stability during training. This can be owed to the fact that the guiding model and the trained
\begin{wraptable}{r}{0.6\textwidth}
\vspace*{-2pt} 
\caption{Comparisons across diffusion backbones and acceleration methods on ImageNet 256×256 using CFG. $\downarrow$ and $\uparrow$ indicate whether lower or higher values are better, respectively.}
\label{tab:cfg_compare}

\resizebox{\linewidth}{!}{
\begin{tabular}{@{}llrrrrr@{}}
\toprule
\textbf{Model} & \textbf{Epochs} & \textbf{FID$\downarrow$} & \textbf{sFID$\downarrow$} & \textbf{IS$\uparrow$} & \textbf{Pre.$\uparrow$} & \textbf{Rec.$\uparrow$} \\
\midrule
\multicolumn{7}{l}{\textit{U-Net}} \\
LDM-4               & 200  & 3.60 & -    & 247.7 & 0.87 & 0.48 \\
\midrule
\multicolumn{7}{l}{\textit{Transformer + U-Net hybrid}} \\
U-ViT-H/2           & 240  & 2.29 & 5.68 & 263.9 & 0.82 & 0.57 \\
MDTv2-XL/2         & 1080 & 1.58 & 4.52 & \textbf{314.7} & 0.79 & 0.65 \\
\midrule
\multicolumn{7}{l}{\textit{Transformer}} \\
MaskDiT             & 1600 & 2.28 & 5.67 & 276.6 & 0.80 & 0.61 \\
SD-DiT              & 480  & 3.23 & -    & -     & -    & -    \\
DiT-XL/2            & 1400 & 2.27 & 4.60 & 278.2 & \textbf{0.83} & 0.57 \\
SiT-XL/2            & 1400 & 2.05 & 4.50 & 270.3 & 0.82 & 0.59
\\
LightningDiT-XL/1   & 800 & 1.35 & 4.15 & 295.3 & 0.79 & 0.65 \\
\midrule
\multicolumn{7}{l}{\textit{Training Acceleration}} \\
SiT-XL/2&  &  & &  & &  \\
\quad + REPA  & {800}    & 1.42 & {4.70} & {305.7} & 0.80 & {0.65} \\
\quad + Disperse Loss  & $\ge$1200 & 1.97 & - & - & - & -\\
\quad + SRA  & 800 &  1.58 & 4.65 & 311.4 & 0.80 & 0.63 \\
\quad + LayerSync  & 800 & 1.89 & - & 265.3 & 0.81 & 0.60 \\
 \rowcolor{lightpurple} \quad + Ours  & \textbf{400} &  1.44 & 4.85 & 311.3 & 0.79 & \textbf{0.66} \\
\midrule
 LightningDiT-XL/1&  &  & &  & &  \\

  \rowcolor{lightpurple} \quad+ Ours & \textbf{400}   & \textbf{1.25} & \textbf{4.11} & 303.9 & 0.78 & \textbf{0.66} \\

\bottomrule
\end{tabular}
}
\vspace{-7mm}
\end{wraptable}
model in our method share a similar architecture and learning process, so that the shallow layers are easier to learn the semantic and structural information.

\noindent\textbf{Comparison of Diffusion Models using CFG}.
We quantitatively compare Self-Transcendence against recent latent diffusion models using two backbones, SiT-XL/2 and LightningDiT-XL, in the Table \ref{tab:cfg_compare}. 
For the SiT-XL/2 backbone,
our method shows significant performance improvement compared with the other 
self-contained methods (SRA and LayerSync), and achieves similar performance 
to REPA at 800 epochs in most metrics using just 400 epochs. 
Specifically, the
baseline SiT-XL achieves a FID of 2.05 and IS of 270.3 after 1400 epochs. 
Incorporating REPA \cite{repa} significantly improves the generation performance, reducing FID to 1.42 and increasing IS to 305.7, with 800 epochs. Our Self-Transcendence framework shows stronger generation acceleration than REPA, achieving an FID of 1.44, IS of 311.3 in only 400 epochs. In addition, leveraging the semantically rich latent space of VAVAE \cite{vavae}, our method outperforms all the compared diffusion models using only 400 epochs in the LightningDiT-XL backbone, with a state-of-art FID result of 1.25.

\subsection{Ablation Study}
\label{sec:ablation}
We conduct a series of ablation studies to investigate the effectiveness of each component of Self-Transcendence and the selection of hyperparameters. Note that the metrics in the study are calculated on 10,000 samples.
\noindent\textbf{Effectiveness of Each Component}.
We ablate Self-Transcendence by removing each key component: VAE structure guidance and self-guided representation. Results are in Table~\ref{tab:ablation_com}.

\begin{wraptable}{r}{0.6\textwidth}
\vspace*{-2pt} 
\caption{Ablation studies. All SiT/B-2 models are trained with 80 epochs.}
\label{tab:ablation_com}

\footnotesize

\resizebox{\linewidth}{!}{
\begin{tabular}{lcc|ccc}
\toprule
\textbf{Methods} & \textbf{VAE structure guid.} & \textbf{Self-guided rep.} & \textbf{FID$\downarrow$}  & \textbf{IS$\uparrow$}   \\
\midrule
SiT-B/2 & {\color{red} \texttimes} & {\color{red} \texttimes} & 38.60 & 41.95   \\
V1       & {\color{blue} \checkmark} & {\color{red} \texttimes} & 32.20 & 52.38   \\
V2       & {\color{red} \texttimes} & {\color{blue} \checkmark} & 25.21& 63.83   \\
 \rowcolor{lightpurple} Ours  & {\color{blue} \checkmark} & {\color{blue} \checkmark} & \textbf{22.91} & \textbf{70.37}   \\
\bottomrule
\end{tabular}}


  \vspace*{-2pt} 
\end{wraptable}

In variant V1, the model is trained only with the VAE structure guidance loss and the standard diffusion loss. Since VAE features are weak in semantics, removing self-guidance slows convergence and hurts performance.
In V2, we use the model trained solely with diffusion loss as the guiding model in the self-guided stage. Without VAE feature guidance, the teacher fails to learn structurally meaningful signals within 40 epochs.
Overall, the full model equipped with both components achieves the best performance, confirming the complementary roles of structure guidance and self-guided representation.

\noindent\textbf{Hyperparameters}.
We ablate the guiding model, loss weight $\lambda_{guide}$, guidance scale $\omega$, and the choice of guiding/guided layers, as shown in the Table \ref{tab:layers-scale}. The top row is our default setting (layer 6 guided by layer 8 with $\omega=30.0$).

\textit{(1) Guiding and guided layers}.
We evaluate different layer pairs as shown in Table \ref{tab:layers-scale}. `$m \rightarrow n$' means that the $m^{th}$ layer of the guiding model is used to supervise the $n^{th}$ layer of the current model. 
Guiding shallow layers (\eg, 8$\rightarrow$6 and 8$\rightarrow$4) consistently outperforms guiding deep layers (8$\rightarrow$8), likely because constraining layers close to the output limits adaptation.
Guiding the same layer depth (6$\rightarrow$6) results in noticeably worse performance, suggesting a small abstraction gap is needed. 
Using very deep guiding layers (10$\rightarrow$6) slightly under-performs 8$\rightarrow$6, indicating overly semantic features are less effective for shallow layers (as also noted in REPA \cite{repa}).
Overall, good guidance requires features that are semantically strong yet well-aligned with the target layer.

\textit{(2) Guidance scale}.
We vary the guidance scale in the self-guided stage from 1.0 to 60.0 (Table middle). A small scale (1.0) greatly hurts performance, showing that weak guidance cannot transfer semantics well. A larger scale (45.0) slightly
improves IS but does not outperform the default setting. An extreme scale (60.0)
degrades both FID and IS, likely due to instability or overfitting to intermediate features \cite{kd1, kd2}. Overall, a
\begin{wraptable}{r}{0.6\textwidth}
\vspace*{-2pt} 
\caption{{Ablation studies on the guidance scale, guiding and guided layers, the selection of the guiding model, and the loss weight $\lambda_{guide}$. `$m \rightarrow n$' means the $n^{th}$ layer of DiT model is guided by the $m^{th}$ layer of guiding model.}}
\label{tab:layers-scale}


\scriptsize
\resizebox{\linewidth}{!}{
\begin{tabular}{cccc|cc}
\toprule
\makecell{Block \\ Layers} & \makecell{Guidance \\Scale}  & \makecell{Guiding model, \\Training iters}& $\lambda_{guide}$ & FID$\downarrow$ & {IS$\uparrow$}  \\
\midrule
\rowcolor{lightpurple} 8 $\rightarrow$ 6    & 30.0  & $M_{align.}$, 200K & 0.5 & 22.91 & 70.37  \\
\midrule
8 $\rightarrow$ 4    & 30.0 & $M_{align.}$, 200K & 0.5& 23.43 & 70.25  \\
8 $\rightarrow$ 8   & 30.0 & $M_{align.}$, 200K & 0.5& 24.29 & 66.88  \\
\midrule
6 $\rightarrow$ 6    & 30.0 & $M_{align.}$, 200K &  0.5& 24.37 & 67.89  \\
10 $\rightarrow$ 6   & 30.0 & $M_{align.}$, 200K &  0.5& 24.05 & 67.94  \\
\midrule
8 $\rightarrow$ 6    & 1.0 & $M_{align.}$, 200K &  0.5& 29.30 & 55.68  \\
8 $\rightarrow$ 6    & 15.0 & $M_{align.}$, 200K &  0.5& 24.26 & 68.36  \\
8 $\rightarrow$ 6    & 45.0 & $M_{align.}$, 200K &  0.5& 23.01 & 70.69  \\
8 $\rightarrow$ 6    & 60.0 & $M_{align.}$, 200K &  0.5& 23.57 & 68.20  \\
\midrule
8 $\rightarrow$ 6    & 30.0 & $M_{align.}$, 50K &  0.5& 28.32 & 57.22  \\
8 $\rightarrow$ 6    & 30.0 & $M_{align.}$, 100K &  0.5& 25.20 & 64.05 \\
8 $\rightarrow$ 6    & 30.0 & $M_{align.}$, 300K &  0.5& 23.05 & 70.32 \\
8 $\rightarrow$ 6    & 30.0 & $M_{ori}$, 200K &  0.5& 25.21 & 63.83 \\
8 $\rightarrow$ 6    & 30.0 & $M_{repa}$, 200K &  0.5 & 23.40 &70.81  \\
\midrule
8 $\rightarrow$ 6    & 30.0 & $M_{align.}$, 200K & 0.1& 24.28 & 67.50 \\
8 $\rightarrow$ 6    & 30.0 & $M_{align.}$, 200K & 0.3& 23.08 & 71.41  \\
8 $\rightarrow$ 6    & 30.0 & $M_{align.}$, 200K & 0.7& 23.27 & 70.12  \\
8 $\rightarrow$ 6    & 30.0 & $M_{align.}$, 200K & 1.0& 23.24 & 68.97  \\
\bottomrule
\end{tabular}}

  \vspace*{-2pt} 
\end{wraptable}
moderate scale (\textit{ie.}, 30.0) best balances gains.

\textit{(3) Guiding model}.
Thirdly, we study the effect of the guiding model. Models trained with VAE structure guidance, vanilla diffusion loss, and REPA are 
denoted as $M_{align}$,  $M_{ori}$, and $M_{repa}$, respectively.
\textbf{Training length.} We train $M_{align}$ for 50K–300K iterations. Better-trained teachers improve FID (\textit{e.g.}, 28.32 at 50K vs. 22.91 at 200K), but 300K brings no further gain (FID 23.05).
This implies that around 200K steps, the intermediate features of the guiding model become semantically rich enough to offer useful supervision. 
With further training, its internal representations may shift from the current model. This is similar to the representation drift problem observed in knowledge distillation \cite{kd1, kd2}, where a too-strong teacher may misguide the student.
\textbf{Teacher type.} we compare different types of guiding models. Using a model trained with standard diffusion loss ($M_{ori}$) results in worse performance, showing that the lack of VAE structure guidance reduces guiding quality. Meanwhile, using a model trained with REPA ($M_{repa}$) with a stronger performance also achieves worse performance than ours, suggesting our carefully designed guiding model $M_{align.}$ provides both the more effective structural and semantic priors.

\textit{(4) Loss weight}.
We ablate the loss weights in Self-Transcendence. Training uses diffusion loss  $\mathcal{L}_{diff}$ and the self-guided loss $\mathcal{L}_{guide}$, with $\lambda_{diff}=1.0$ and $\lambda_{guide} \in [0.1,1.0]$ (Table \ref{tab:layers-scale}). The best result is at $\lambda_{guide}=0.5$ (FID 22.91, IS 70.37). Too small $\mathcal{L}_{guide}$ weakens guidance ($\lambda_{guide}=0.1$: FID 24.28), while too large $\lambda_{guide}$ hurts performance, likely due to over-regularization ($\lambda_{guide}=1.0$: FID 23.24). 
These results confirm the importance of balancing the two loss terms. A moderate guidance weight (\eg, 0.5) provides optimal control without overwhelming the primary diffusion objective. Therefore, in our final model, we choose $\lambda_{guide}=0.5$ as the default setting.

\begin{table}[t]
\scriptsize
\caption{Training cost comparison on SiT-B/2 with batch size 256 on ImageNet $256 \times 256$ using 8$\times$A800 GPUs.}
\label{tab:training_cost}
\resizebox{\linewidth}{!}{
\begin{tabular}{lccc}
\toprule
Method & Speed (iters/s)$\uparrow$ & Total time (h)$\downarrow$ & Peak memory (GB/card)$\downarrow$ \\
\midrule
Vanilla (SiT) \cite{sit} 
& 9.28 & 58.87 & 9.03 \\
\midrule
REPA \cite{repa} 
& 8.49 & 65.44 & 10.39 \\
\midrule
 \rowcolor{lightpurple} Self-Transcendence 
& \makecell[l]{Stage-1: 6.69 (first 50K iters) \\ Stage-2: 9.28 (remaining 1950K iters)} 
& 60.45 
& \makecell[l]{12.34 (during first 50K iters) \\ 9.03 (after early stop)} \\
\bottomrule
\vspace{-7mm}
\end{tabular}}
\end{table}

\subsection{Training Cost Comparison}
We compare the training cost of Self-Transcendence, REPA \cite{repa}, and the vanilla model \cite{sit} in terms of both time and GPU memory in Table \ref{tab:training_cost}.
All models are trained with batch size 256.
Using SiT-B/2 as an example, we train for 
400 epochs (2,000K iterations) on 8 A800 GPUs.
The vanilla model runs at 9.28 iters/s, taking about 58.87 hours in total.
Its peak GPU memory is about 9.03~GB per card.
With REPA, the speed drops to 8.49 iters/s, resulting in 65.44 hours.
REPA relies on a pretrained DINOv2 model \cite{dinov2}, which brings extra cost
and increases the peak memory to 10.39~GB.
For Self-Transcendence, the first 50K iterations run at 6.69 iters/s, and the remaining 1950K iterations run much faster, reaching 9.28 iters/s, leading to 60.45 hours in total.
In terms of memory, our method only adds a guiding model in the early stage, whose peak memory is 12.34~GB during the first 50K iterations, and then returns to the vanilla level (9.03~GB) after early stop.

Overall, our method has a comparable training time to the vanilla model and is faster than REPA.
It also has a smaller memory overhead than REPA, and keeps the long-run memory cost close to vanilla.
Moreover, training our guiding model takes only 6.39 hours (200K iterations at 8.70 iters/s), which is substantially cheaper than pretraining a large teacher such as DINOv2.

\subsection{Understanding Capability.}

\begin{wraptable}{r}{0.6\textwidth}
\vspace*{-2pt} 
\caption{Comparisons on linear probing on ImageNet classification using different DiT layers features.}
\label{tab:capacity}
\tiny
\resizebox{\linewidth}{!}{
\begin{tabular}
{p{1.1cm}|m{0.7cm}m{0.7cm}m{0.7cm}m{0.7cm}m{0.7cm}m{0.7cm}} 
\hline
 Layers & 4    & 8   & 12  & 16  & 20  & 24   \\ \hline
SiT-XL/2   & 11.19\%   & 11.87\%   & 16.25\%   & 29.49\%   & 42.15\%   & 39.13\% \\ \hline
+ REPA  & 26.47\%   & 51.87\%   & 52.07\%   & 49.76\%   & 47.98\%   & 44.52\% \\ 
\hline
 \rowcolor{lightpurple} + Ours   & 25.06\%   & 41.18\%   & 49.13\%   & 50.00\%   & 47.46\%   & 43.52\% \\ 
\hline
\end{tabular}}
  \vspace*{-2pt} 
\end{wraptable}

To show the improvement on understanding capability, we evaluate the features at $t=0.5$ by linear probing 
on ImageNet classification in Table \ref{tab:capacity}. Self-transcendence clearly improves vanilla SiT-XL with enhanced discriminative capability, though it stays below REPA, which uses a DINO-pretrained teacher \cite{dinov2}. Notably, stronger linear separability does not necessarily mean better generation: our guiding features share the same diffusion setting as the student and are thus better aligned with the generative objective, explaining our superior generation.

\section{Conclusion and Limitation}
\label{sec:conclusion}
\vspace{-0.2em}
 
We proposed \textbf{Self-Transcendence}, a simple yet effective self-guided training framework to accelerate the training of diffusion transformers (DiTs). Unlike previous approaches that relied on external models for semantic supervision, our method was entirely self-contained by leveraging the model’s own internal features to guide its training. We designed a two-stage pipeline, where we first aligned shallow-layer features with VAE latents to provide stable early supervision and then applied classifier-free guidance to enhance the semantic expressiveness of intermediate features. The obtained features were used to guide a new DiT training. Extensive experiments demonstrated that our method achieved comparable or even superior performance to externally guided methods such as REPA. Our findings highlighted the untapped potential of internal representations in DiT models and provided a new direction for self-supervised acceleration.

\noindent\textbf{Limitations}.  
While our method eliminates the need for external models, the quality of internal guidance is still upper-bounded by the model's own capacity. Secondly, our method introduces additional hyperparameters, which may require tuning across tasks. Finally, as in previous works \cite{wang2025ddt, disperse, LayerSync, vavae}, our approach is evaluated on image generation benchmarks; its applicability to other generative modalities (\eg, text-to-video generation and text-to-3D generation) deserves to be explored in future work.

{\small
\bibliography{ref}
}

\appendix

\clearpage
\setcounter{table}{0}
\setcounter{equation}{0}
\setcounter{figure}{0}
\renewcommand{\thetable}{\thesection.\arabic{table}}
\renewcommand{\theequation}{\thesection.\arabic{equation}}
\renewcommand{\thefigure}{\thesection.\arabic{figure}}
\begin{center}
    {\LARGE\bfseries Appendix}
\end{center}

\noindent The following materials are provided in this Appendix:
\begin{itemize}
    \item Comparisons between two-stage training and a one-stage variant (see Sec. 3.1 of the main paper).
    \item The MLP architecture used for feature alignment (see Sec. 3.3 of the main paper).
    \item The early stop strategy for the self-guided loss (see Sec. 4.1 of the main paper).
    \item ImageNet \(512 \times 512\) experiments (see Sec. 4.2 of the main paper).
    \item Feature map visualizations via PCA.
\end{itemize}

\section{Comparisons of Two-Stage and One-Stage Training}
\label{sec: one-stage}
To further validate the benefit of the two-stage training design, we compare it with a one-stage variant. In the one-stage setting, we jointly optimize the diffusion loss, the VAE-based structural guidance loss, and the self-guided loss. During training, we use the updated Exponential Moving Average (EMA) model as the guiding model and apply the self-guided representation at its intermediate features to guide the optimization of the online model. All experiments are performed using the SiT-B/2 backbone, and 50,000 samples are used for evaluation.
Results are shown in Table \ref{tab:one-stage}. We see that the one-stage training performs worse than our two-stage method. This is because early training is critical, but the EMA model is still weak at the beginning and cannot provide reliable guidance. In contrast, our two-stage training first builds a strong initialization in Stage 1, and then applies self-guided representations to enhance the semantic expression in Stage 2 when the model becomes sufficiently strong.

\begin{table}[ht]
\centering
\scriptsize
\caption{{Comparisons of two-stage and one-stage training on the SiT-B/2 backbone. $\downarrow$ and $\uparrow$ indicate whether lower or higher values are better, respectively.}}
\begin{tabular}{lcccccc}
\toprule
\textbf{Model} & \textbf{Epochs} & \textbf{FID$\downarrow$} & \textbf{sFID$\downarrow$} & \textbf{IS$\uparrow$} & \textbf{Pre.$\uparrow$} & \textbf{Rec.$\uparrow$} \\
\midrule
SiT-B/2 & 80 & 36.14 &  6.68 &41.75 & 0.5211 & \textbf{0.6337}\\
\midrule
\quad + One-stage & 80 & 34.50 & 6.79 & 43.46 & 0.5355 & 0.6308 \\
\midrule
\rowcolor{lightpurple} \quad + Two-stage & 80 & \textbf{20.49} &\textbf{6.68} & \textbf{70.57} &  \textbf{0.6161} & 0.6321 \\
\bottomrule
\end{tabular}

\label{tab:one-stage}
\end{table}

\section{MLP Architecture}
\label{sec: mlp}
We use a simple multilayer perceptron (MLP) to map the model's hidden state to the target representation space.  
The architecture follows REPA \cite{repa} for a fair comparison, which aligns with the DINO representation \cite{dinov2}.  
Specifically, the MLP consists of three linear layers with two SiLU activations in between.  
The input is first projected to an intermediate dimension, then transformed twice before outputting the final representation.

\begin{figure*}[b]
	\centering 
	\includegraphics[scale=0.5]{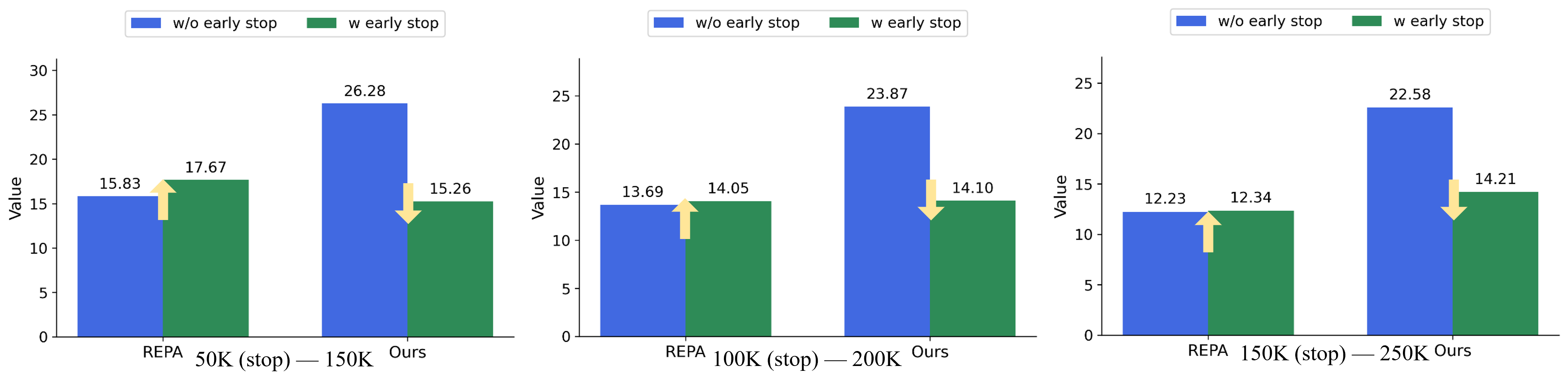}
	\vspace{-2mm}
	\caption{FID-10K scores with and without early stopping at various training stages of REPA and our Self-Transcendence method. For example, `50K (stop) — 150K' means that the alignment loss is not optimized after 50K iterations, and only the diffusion loss is used to optimize to 150K iterations. One can see that our Self-Transcendence method benefits from early stopping, achieving better FID scores, while REPA’s performance degrades when the early stop strategy is applied.
    }
	\label{earlystop}
\vspace{-2mm}
\end{figure*}

\section{Early Stop Strategy}
\label{sec: early-stop}
We investigate the effect of early stopping on our method during training. Specifically, we stop optimizing the self-guided loss after a certain number of iterations (\eg, 50K) and continue training using only the diffusion loss for 100K more iterations (\eg, 150K). This strategy is indicated by \textit{50K (stop) — 150K}. All experiments are performed using the SiT-XL/2 backbone, and 10,000 samples are used for evaluation.
As shown in Fig. \ref{earlystop}, our method benefits from early stopping, achieving better FID-10K scores; however, REPA's performance degrades when the alignment loss is stopped early.
One possible reason is that over-training the shallow layers with semantic alignment may destabilize the training of deeper layers and hinder the modeling of joint data distribution. In contrast, our method can take advantage of early stopping, leading to a lower total computational cost.

\begin{table}[ht]
\centering
\footnotesize
\caption{{Comparisons across diffusion backbones and acceleration methods on ImageNet 512×512 using CFG. $\downarrow$ and $\uparrow$ indicate whether lower or higher values are better, respectively.}}
\begin{tabular}{lcccccc}
\toprule
\textbf{Model} & \textbf{Epochs} & \textbf{FID$\downarrow$} & \textbf{sFID$\downarrow$} & \textbf{IS$\uparrow$} & \textbf{Pre.$\uparrow$} & \textbf{Rec.$\uparrow$} \\
\midrule
\multicolumn{7}{l}{\textit{Pixel diffusion}} \\
VDM++ & -- & 2.65 & -- & 278.1 & -- & -- \\
\midrule
\multicolumn{7}{l}{\textit{Latent diffusion, Transformer}} \\
 MaskDiT & 800 & 2.50 & 5.10 & 256.3 & 0.83 & 0.56 \\
 DiT-XL/2 & 600 & 3.04 & 5.02 & 240.8 & \textbf{0.84} & 0.54 \\
 SiT-XL/2 & 600 & 2.62 & 4.18 & 252.2 & \textbf{0.84} & 0.57 \\
\quad + REPA & 200 & 2.08 & 4.19 & 274.6 & 0.83 & 0.58 \\
\rowcolor{lightpurple} \quad + Ours & 100 & 2.00 & \textbf{4.11} & 265.0 & 0.83 & 0.58 \\
\rowcolor{lightpurple} \quad + Ours & 200 & \textbf{1.76} & 4.16 & \textbf{286.6} & 0.82 & \textbf{0.62} \\
\bottomrule
\end{tabular}

\label{tab:diffusion-512}
\end{table}

\begin{figure*}[b]
	\centering 
	\includegraphics[scale=0.7]{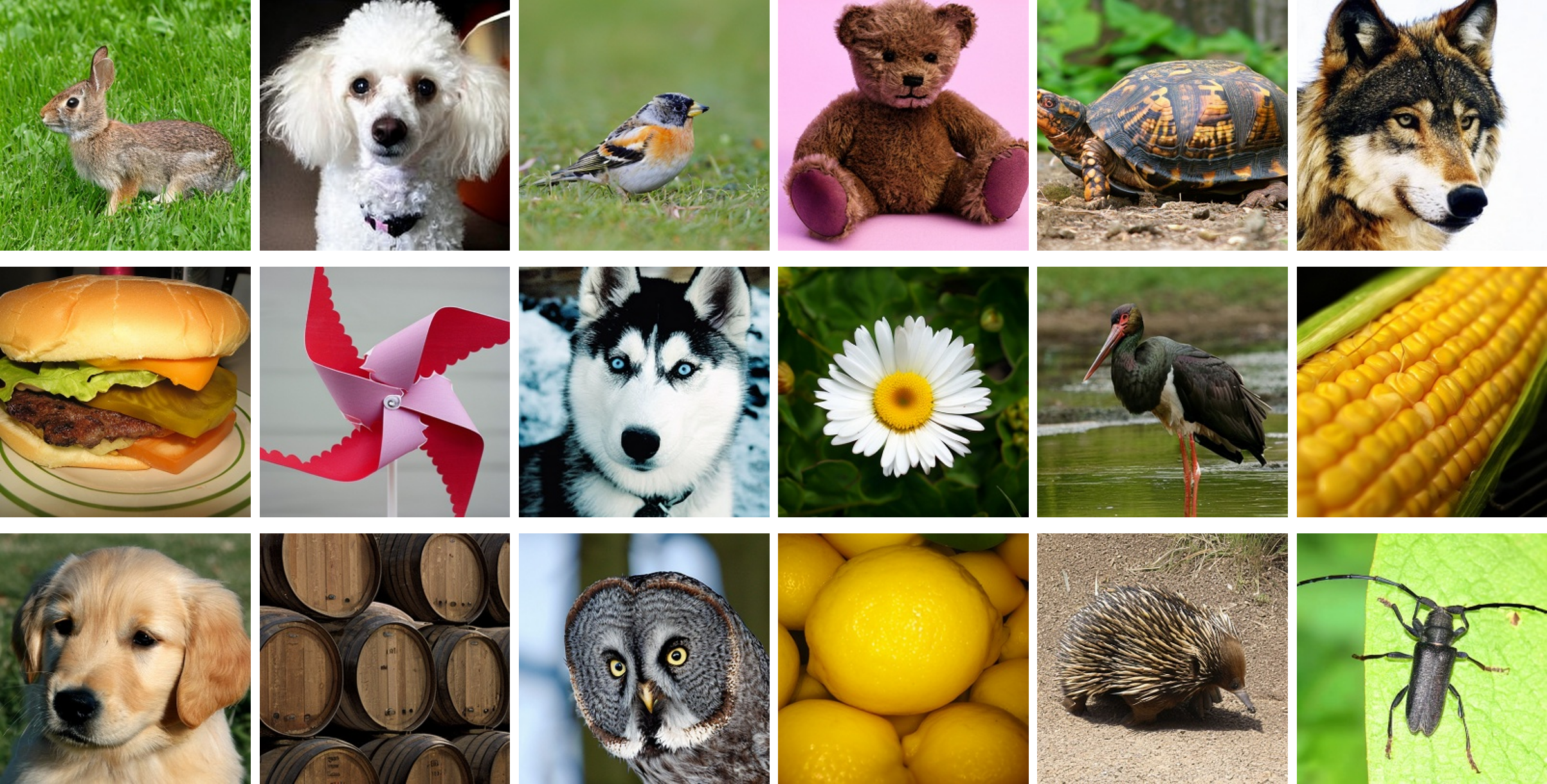}
	\vspace{-2mm}
	\caption{Examples of generated images on ImageNet $512 \times 512$ of our proposed Self-Transcendence method. We use classifier-free guidance with 4.0 scale.
    }
	\label{generated512}
\vspace{-2mm}
\end{figure*}

\section{ImageNet $512 \times 512$ Experiment}
\label{sec: 512}
To further evaluate the robustness and scalability of our method in higher-resolution scenarios, we conduct experiments on ImageNet at $512 \times 512$ resolution. This setting poses greater challenges for both semantic alignment and visual fidelity due to the increased spatial complexity and richer details.
We employ the same training pipeline as REPA \cite{repa} for a fair comparison.
As shown in Table \ref{tab:diffusion-512}, the baseline SiT-XL/2 model \cite{sit} exhibits limited performance at this resolution, with a FID of 2.62 and IS of 252.2 after 600 training epochs. Incorporating REPA \cite{repa} significantly improves the generation performance, reducing FID to 2.08 and increasing IS to 274.6, with 200 epochs. Our Self-Transcendence framework shows stronger generation acceleration than REPA, achieving an FID of 1.76, IS of 286.6 with the same 200 epochs.
{Actually, even with only 100 epochs, our method can already reach an FID of 2.00 and IS of 265.0, which is close to REPA at 200 epochs, suggesting that our method can obtain strong performance with much fewer training epochs.}
These results demonstrate that the internal features from the DiT model itself can also provide effective guidance. 

We provide qualitative results of SiT-XL/2 on ImageNet at resolutions of 512 using the proposed Self-Transcendence method in Fig. \ref{generated512}, which demonstrates realistic structures and fine textures across diverse semantic categories at higher resolutions. For example, for manmade objects with complex geometry, such as wooden barrel and pinwheel, the generated images by Self-Transcendence exhibit accurate geometry and sharp edges.

\section{Feature Map Visualization}
\label{sec: fea-map}

\begin{figure*}[b]
	\centering 
	\includegraphics[scale=0.6]{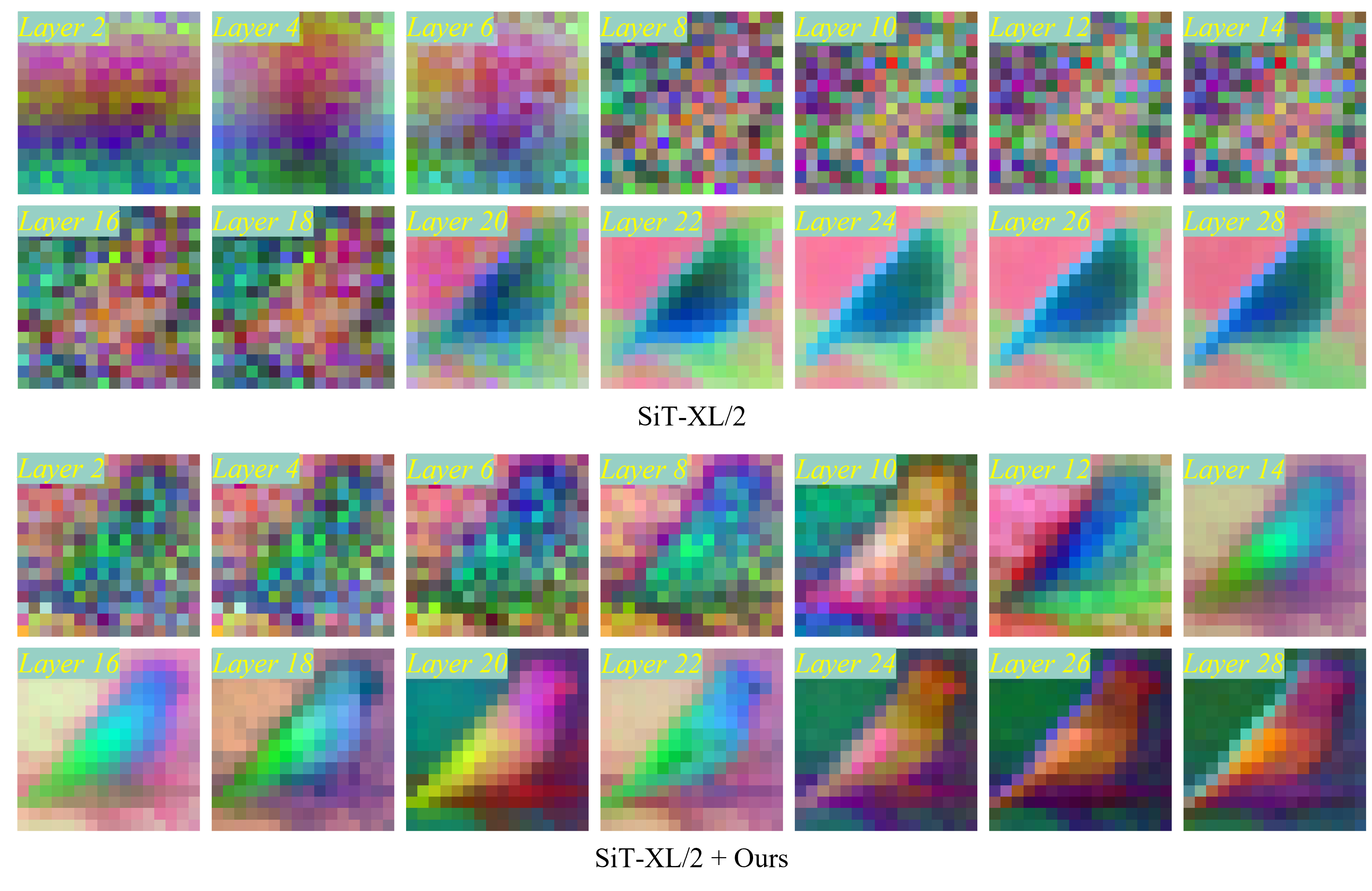}
	\vspace{-2mm}
	\caption{PCA feature visualization from different layers of SiT-XL/2 and SiT-XL/2+Self-Transcendence after 400K iterations and $t=0.7$.
    }
	\label{fea0-7}
\vspace{-2mm}
\end{figure*}

\begin{figure*}
	\centering 
	\includegraphics[scale=0.6]{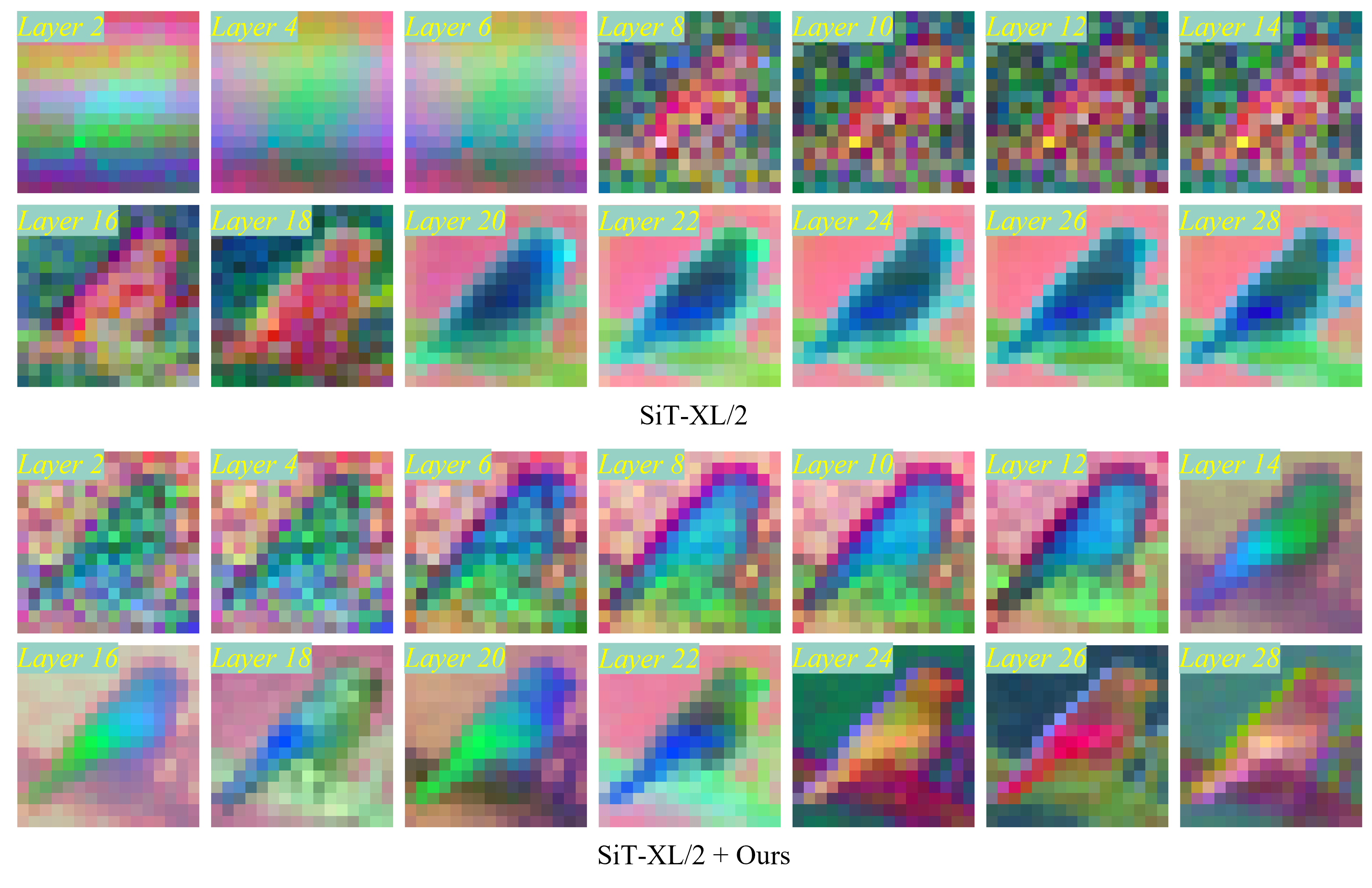}
	\vspace{-2mm}
	\caption{PCA feature visualization from different layers of SiT-XL/2 and SiT-XL/2+Self-Transcendence after 400K iterations and $t=0.5$.
    }
	\label{fea0-5}
\vspace{-2mm}
\end{figure*}

We provide PCA visualizations \cite{abdi2010pca} of feature maps in Fig. \ref{fea0-7} and Fig. \ref{fea0-5} to illustrate the evolution of representation across layers. As can be seen, Self-Transcendence significantly enhances feature organization along different layers, showing more compact and structured representations. In contrast, the vanilla model exhibits more scattered and less coherent patterns, indicating weaker discriminative features.

\end{document}